\documentclass[10pt,twocolumn,letterpaper]{article}

\usepackage{cvpr}      

\usepackage[table, x11names, dvipsnames]{xcolor}

\usepackage{xr-hyper} 

\definecolor{cvprblue}{rgb}{0.21,0.49,0.74}
\usepackage[pagebackref,breaklinks,colorlinks,citecolor=cvprblue]{hyperref}
\usepackage{url}
\usepackage{times}
\usepackage{epsfig}
\usepackage{graphicx}
\usepackage{amsmath}
\usepackage{amssymb}
\usepackage{mwe}
\usepackage{placeins}
\usepackage{booktabs}
\usepackage{comment}
\usepackage{xspace}
\usepackage{multirow}

\usepackage[normalem]{ulem} 

\usepackage[accsupp]{axessibility} 

\graphicspath{{figs/}}

\usepackage{amsmath,amsfonts,bm}

\def\eqref#1{equation~\ref{#1}}

\def\1{\bm{1}}

\DeclareMathAlphabet{\mathsfit}{\encodingdefault}{\sfdefault}{m}{sl}
\SetMathAlphabet{\mathsfit}{bold}{\encodingdefault}{\sfdefault}{bx}{n}

\newcommand{\name}{WinSyn\xspace} 

\newcommand{\imgcount}{75,739\xspace}
\newcommand{\cropcount}{89,318\xspace}
\newcommand{\syncount}{21,290\xspace} 

\newcommand{\labelcount}{9,002\xspace}

\newcommand{\testsize}{4,906\xspace}

\newcommand{\paramcount}{21,735\xspace} 

\newcommand{\variationcount}{64\xspace}
\newcommand{\totalrenders}{156,903\xspace} 

\newcommand{\basemiou}{32.58\xspace} 
\newcommand{\realmiou}{58.69\xspace}

\newcommand{\facades}{fa\c{c}ades\xspace}
\newcommand{\facade}{fa\c{c}ade\xspace} 
\def\paperID{11645} 
\def\confName{CVPR}
\def\confYear{2024}

\title{\name: A High Resolution Testbed for Synthetic Data}

\author{Tom Kelly$^1$ \quad John Femiani$^2$ \quad Peter Wonka$^1$ \vspace{0.3em} \\
{\normalsize $^1$KAUST, KSA} \quad
{\normalsize $^2$Miami University, Ohio} \quad
}

\begin{document}
\maketitle

\begin{abstract}

We present {\name}, a unique dataset and testbed for creating high-quality synthetic data with procedural modeling techniques.
The dataset contains high-resolution photographs of windows, selected from locations around the world, with \cropcount individual window crops showcasing diverse geometric and material characteristics. 
We evaluate a procedural model by training semantic segmentation networks on both synthetic and real images and then comparing their performances on a shared test set of real images. Specifically, we measure the difference in mean Intersection over Union (mIoU) and determine the effective number of real images to match synthetic data's training performance.
We design a baseline procedural model as a benchmark and provide \syncount synthetically generated images.
By tuning the procedural model, key factors are identified which significantly influence the model's fidelity in replicating real-world scenarios.
Importantly, we highlight the challenge of procedural modeling using current techniques, especially in their ability to replicate the spatial semantics of real-world scenarios.
This insight is critical because of the potential of procedural models to bridge to hidden scene aspects such as depth, reflectivity, material properties, and lighting conditions.




\end{abstract}

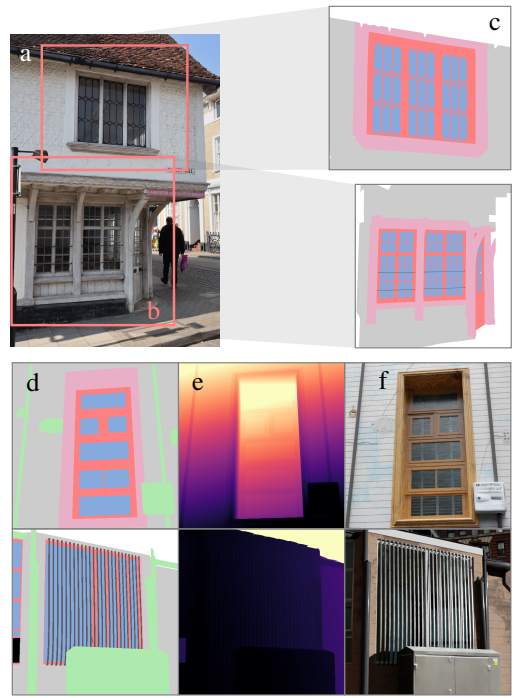
\begin{figure}[!b]
    \centering
    \def\svgwidth{0.8\linewidth}
\begingroup%
  \makeatletter%
  \providecommand\color[2][]{%
    \errmessage{(Inkscape) Color is used for the text in Inkscape, but the package 'color.sty' is not loaded}%
    \renewcommand\color[2][]{}%
  }%
  \providecommand\transparent[1]{%
    \errmessage{(Inkscape) Transparency is used (non-zero) for the text in Inkscape, but the package 'transparent.sty' is not loaded}%
    \renewcommand\transparent[1]{}%
  }%
  \providecommand\rotatebox[2]{#2}%
  \newcommand*\fsize{\dimexpr\f@size pt\relax}%
  \newcommand*\lineheight[1]{\fontsize{\fsize}{#1\fsize}\selectfont}%
  \ifx\svgwidth\undefined%
    \setlength{\unitlength}{325.46806756bp}%
    \ifx\svgscale\undefined%
      \relax%
    \else%
      \setlength{\unitlength}{\unitlength * \real{\svgscale}}%
    \fi%
  \else%
    \setlength{\unitlength}{\svgwidth}%
  \fi%
  \global\let\svgwidth\undefined%
  \global\let\svgscale\undefined%
  \makeatother%
  \begin{picture}(1,1.37639211)%
    \lineheight{1}%
    \setlength\tabcolsep{0pt}%
    \put(0,0){\includegraphics[width=\unitlength,page=1]{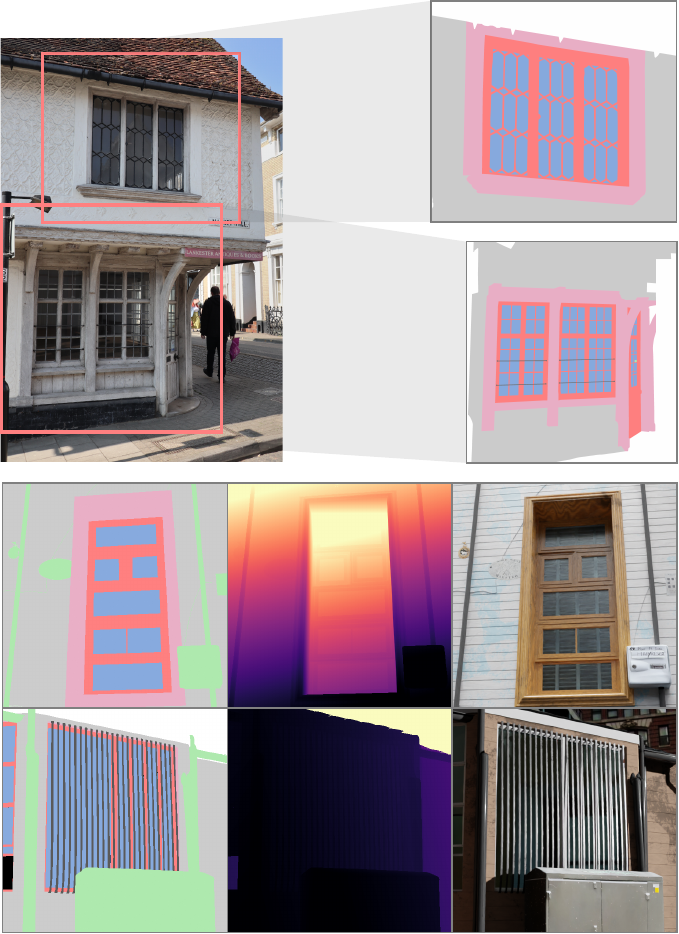}}%
    \put(0.28635176,0.75227672){\color[rgb]{1,0.50196078,0.50196078}\makebox(0,0)[t]{\lineheight{1.25}\smash{\begin{tabular}[t]{c}b\end{tabular}}}}%
    \put(0.0308762,1.26548026){\color[rgb]{1,1,1}\makebox(0,0)[t]{\lineheight{1.25}\smash{\begin{tabular}[t]{c}a\end{tabular}}}}%
    \put(0.96626928,1.32891036){\color[rgb]{0,0,0}\makebox(0,0)[t]{\lineheight{1.25}\smash{\begin{tabular}[t]{c}c\end{tabular}}}}%
    \put(0.74386576,0.61103274){\color[rgb]{0,0,0}\makebox(0,0)[t]{\lineheight{1.25}\smash{\begin{tabular}[t]{c}f\end{tabular}}}}%
    \put(0.37424369,0.61044727){\color[rgb]{0,0,0}\makebox(0,0)[t]{\lineheight{1.25}\smash{\begin{tabular}[t]{c}e\end{tabular}}}}%
    \put(0.04555506,0.60817267){\color[rgb]{0,0,0}\makebox(0,0)[t]{\lineheight{1.25}\smash{\begin{tabular}[t]{c}d\end{tabular}}}}%
  \end{picture}%
\endgroup%

    \caption{Photographers in 28 geographic regions captured real-world photos (a) of windows that are cropped (b) to single windows, which are then labeled (c). Synthetic windows are rendered giving color (f) and labels(d), while other passes such as depth (e) are also possible.}
    \label{fig:overview}
\end{figure}

\section{Introduction}

Larger and more sophisticated machine learning models demand an ever-increasing supply of data, particularly in tasks where manual annotation is challenging, such as depth estimation, reflectance estimation, or full 3D reconstruction. One solution is to use procedural models to generate synthetic training data.  However, creating procedural models that accurately reflect the domain of real images (i.e. closing the domain gap) remains an open problem. Despite the visual realism of many synthetic scenes, their effectiveness in machine-learning applications often falls short. In this work we do not close the gap -- but
present an accessible pair of real and synthetic datasets, with annotations, in which this domain gap may be studied without massive resources.

While the final goal is to tackle complex problems, a straightforward proxy task is initially required to ensure that real-world imagery can be manually annotated and compared to synthetic imagery. We, therefore, propose segmentation as a proxy task. This proxy task helps pinpoint where a procedural model fails to capture the diversity and nuances of real-world scenes.

As our long-term goal is procedural urban modeling, we initially considered street-view images and datasets, such as CityScapes~\cite{Cordts2016Cityscapes}. However, this has several drawbacks. Modeling arbitrary urban scenes realistically requires overwhelming complexity; requiring modeling at least cars, humans, buildings, skies, and vegetation, and each of these poses quite distinct and complex modeling challenges. For example, modeling urban environments for video games is a significant undertaking that often requires over 100 artists for open-world games. This type of effort is unrealistic for procedural modeling research. We propose that it is more promising to identify a simpler subset of images and develop a more constrained procedural model for this subset. This should enable faster iterations and broader participation in procedural modeling research, before scaling up to complete cities.

The decision to study the domain of window images was based on the following design principles for our dataset:
\begin{description}[leftmargin=0pt, labelindent=10pt, topsep=0pt, partopsep=5pt]
\item[{Diversity:}] 
 Our goal is to capture a range of variations in both geometry and materials. We opted against humans due to the complexities in modeling realistic fine geometric details (like hair and beard) and materials (such as skin), and the limited topological variations. Although humans have been used for synthetic data studies as demonstrated in~\cite{wood_fake_2021}, exploration in this domain is complicated because the procedural model was not released and high-fidelity models such as Unreal Metahumans by Epic Games~\cite{EpicGamesMetahumans} tend to prioritize realism over variability. Plants, another option, present challenges in material diversity and are difficult for human annotators to segment. Windows, in contrast, offers a balanced combination of geometric complexity and material variation suitable for our research goals.
 
\item[{High-resolution:}] The images in the dataset should be high resolution to make the dataset useful for the near and medium-term future. This is in contrast to existing architectural datasets, with resolutions typically in the 512-1025 pixel range~\cite{LabelMeFacade-Brust15:ECP, Brust15:ECP, etrims-korc-forstner-tr09, riemenschneider2012irregular}.   While these are not low resolution in typical computer vision terms, architectural features such as windows are often too small to resolve in the images. Although the CityScapes dataset is larger than most, with a resolution of $2048\times1024$ pixels, it often captures architecture from oblique viewpoints limiting the effective resolution.
\item[{Image Rights:}] We would like to have all rights to the images to avoid future copyright problems.
\end{description}

To explore this problem, we introduce a specialized dataset that bridges real-world architectural imagery and procedurally-generated images, with a particular focus on window designs. Our dataset, comprising \cropcount{} windows in \imgcount{} high-resolution (4K to 6K) and RAW images, not only matches the scale of established datasets like CelebA-HQ and FFHQ but also offers a unique niche with an emphasis on architectural details and high resolutions.

Our dataset is designed not just to advance synthetic data generation, but also to facilitate diverse research avenues. Our primary contributions\footnote{Available online \href{https://twak.github.io/winsyn}{https://twak.github.io/winsyn}} are:
\begin{enumerate}
    \item \textbf{Real-World Imagery:} A 4K resolution dataset with \imgcount{} photos of windows from global locations, offering unprecedented detail for architectural research.
    \item \textbf{Hand-Annotated Labels:} Segmentation labels for \labelcount{} images.
    \item \textbf{A Procedural Model:} A novel procedural model for windows, underscoring key design choices for synthetic data generation.
    \item \textbf{Synthetic Data and Labels:} A diverse set of \syncount{} synthetic window images, mirroring features observed in real imagery.
\end{enumerate}

Additionally, we conduct extensive experiments and ablations to understand the impact of various features in our synthetic dataset on segmentation performance.

\section{Related Work}\label{sec:related-work}

Several datasets of architectural imagery have been created, enabling applications such as architectural style classification~\cite{xu2014architectural,chen2021hierarchical,barz2021wikichurches}, building functional use classification~\cite{kang2018building,zhaoBEAUTY}, architectural heritage classification~\cite{llamas2017classification}, landmark identification~\cite{Philbin07}, or urban scene matching~\cite{hauagge2012image}. Notably, datasets that support image synthesis or image segmentation have been instrumental in advancing these fields. For instance, the FaSyn13 dataset~\cite{dai:facade:iccv13} consists of 200 \facade images for texture synthesis, but its size is limited for modern generative models. Similarly, the LSAA dataset~\cite{zhuLSAA} includes 199,723 \facade images and 516,000 cropped window images. However, the resolution of these cropped windows varies, with the majority being less than 100 pixels in the longest dimension.

In comparison, extensive collections like MS COCO~\cite{LinMSCOCO} and LAION~\cite{schuhmann2021laion} offer a broader range but with challenges in resolution and scene variety. For instance, the sheer diversity in LAION's 9.8 million 1K-resolution images presents significant challenges for a single procedural model, highlighting the need for focused datasets. Our dataset addresses this by providing high-resolution window images, offering a specialized resource for detailed architectural analysis and procedural modeling.
 
Our windows-focused dataset, while containing fewer windows than LSAA, stands out as the highest-resolution dataset of window images known to us, with an average resolution of 4,000 pixels per side for cropped windows. This high-resolution focus, particularly in the 4K to 6K range, fills a unique niche in architectural and synthetic-to-real domain transfer research. Our dataset’s specialization in high-resolution architectural elements, especially windows, offers a valuable resource for advancing procedural modeling, emphasizing the importance of detail and precision.

Several architectural image datasets supporting semantic segmentation or \facade parsing exist, though not specifically focused on windows.  The Graz dataset~\cite{riemenschneider2012irregular} contains 50 rectified images, and the eTRIMS datset~\cite{etrims-korc-forstner-tr09} contains 60 non-rectified images. The CMP-Facade dataset~\cite{CMPFacade_Tylecek13} contains 606 images, with about half of them fairly high resolution (1,024 pixels on the long edge) but a limited diversity of image locations. The LabelMe-Facade dataset~\cite{Brust15:ECP} has the largest number of images at 945, with each image varying in size between 512 and 768 pixels on a side. However, these datasets do not have the number of images nor the resolution that are desirable for training the latest computer vision methods, which increasingly require more detailed and high-resolution data. With \labelcount{} labeled images at four times the resolution of these datasets, our proposed dataset of real-world images is an order of magnitude larger. By concentrating exclusively on windows, our dataset offers unprecedented detail and specificity, enabling more precise and effective models for architectural element analysis.

Various authors have attempted to use synthetically generated data to bootstrap performance on real images. This approach seems to work best in domains where the human annotation is not directly feasible, such as reinforcement learning, especially for driving applications~\cite{Dosovitskiy17}, depth or optical flow~\cite{butler2012naturalistic, gaidon2016virtual}, or 6DoF pose estimation for robotic grasping or manipulation~\cite{tyree2022hope, hodan2020bop, kaskman2019homebreweddb}.  

Infinigen~\cite{raistrick2023infinite} is a good example of a procedural model, however, there is no validation of the effectiveness of the model for machine learning tasks. Of particular note is the SynthIA dataset~\cite{ros2016synthia}, a driving dataset built on video game technology specifically designed to support semantic segmentation in urban environments. A very large engineering effort went into this, as well as CARLA~\cite{Dosovitskiy17}, and we believe that reproducing such high-quality synthetic data is out of reach for most academic teams.  Similar to our dataset, SynthIA aims at pushing the envelope to use synthetic data to improve computer vision even for problems where large human-annotated datasets (KITTI~\cite{Menze2015CVPR}, LabelME-Facade~\cite{LabelMeFacade-Brust15:ECP}, Camvid~\cite{ BrostowSFC:ECCV08}) already exist. Similar to our findings, they can get some results from purely synthetic data, but they cannot out-compete even relatively small real-world labeled images, but by combining synthetic and real at a 4:6 ratio they obtain their best results. However, unlike SynthIA, our segmentation challenge is more constrained (only windows) and we think would require fewer resources for academic researchers to develop competing procedural models for synthetic training.  Although buildings are visible in these datasets, they are not focused on architecture. Domain transfer for architecture is challenging due to the amount of variety and complex dependencies between architectural elements, whereas the categories of objects relevant in driving scenes are much more clearly defined. In addition, the high resolution of the images we use makes the synthesis of realistic textures and precise object boundaries critical, and the higher-capacity segmentation models of today vs. 2016 (when SynthIA was published) are more precise but may also be more likely to overfit synthetic data.  Our dataset of real and synthetic imagery is unique as a high-resolution, voluminous dataset and serves as a proving ground for synthetic to real training, image generation, and semantic segmentation tasks. 

\begin{figure}
    \centering
    \def\svgwidth{\linewidth}
    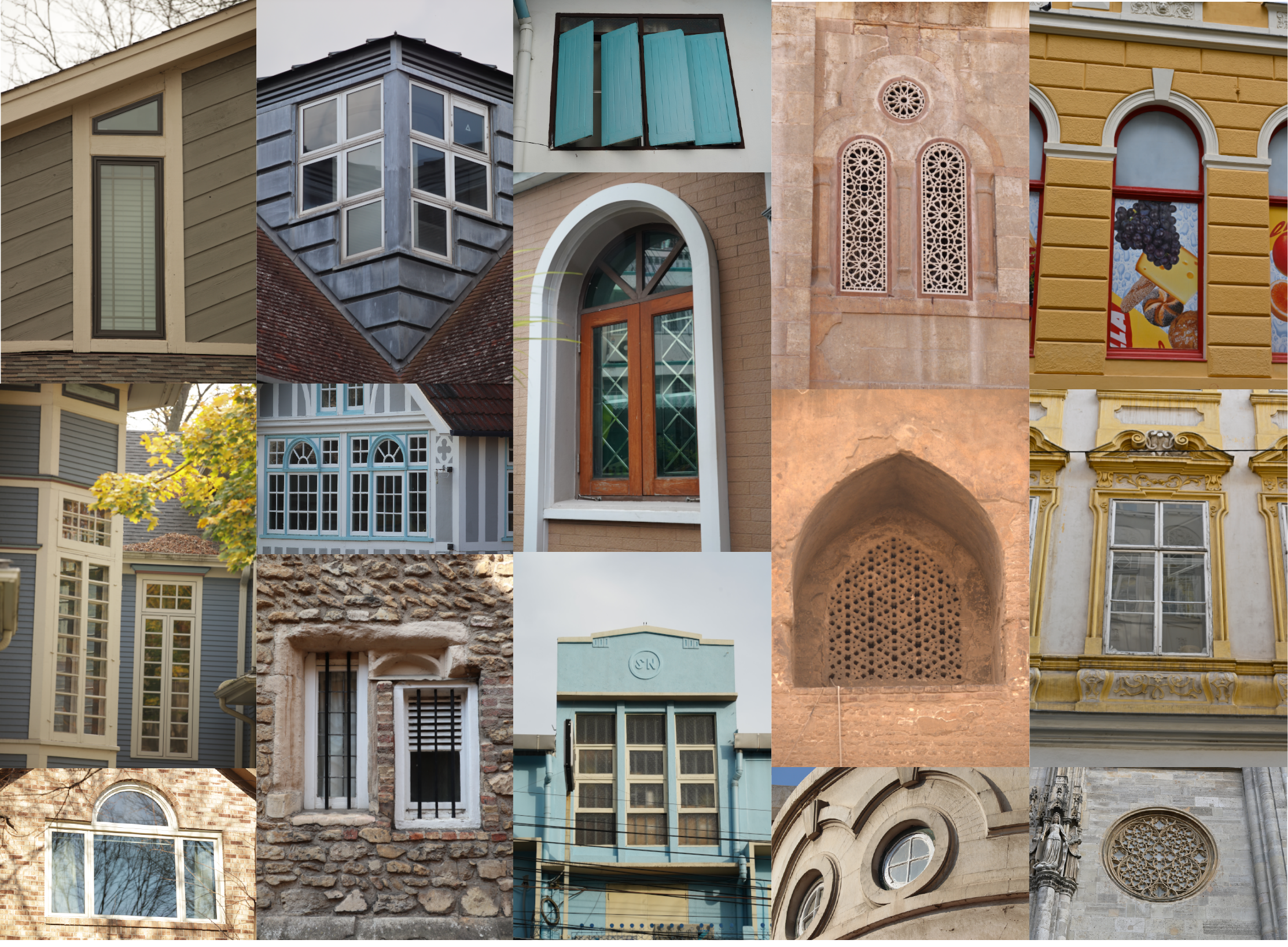
    \caption{Samples from the \imgcount{}~photographs in the dataset. Each column shows a variety of examples of windows from different geographic locations. From left to right: Chicago (USA), Cambridge (UK), Bangkok (Thailand), Cairo (Egypt), and Vienna (Austria). The dataset has a variety of window shapes and architectural styles. }
    \label{fig:photos_twelve}
\end{figure}

To our knowledge, the most successful work on using synthetic data for segmentation is the `Fake It Till You Make It' paper by~\citet{wood_fake_2021}, which reports improvements in segmentation when a U-Net is trained using their synthetic data vs. real image. However, to get these improved results, they used label adaptation, a technique that requires real labeled data. This is counter to our goals for using synthetic data for domains where labels are scarce.
They do not report segmentation results without label adaptation, but they do show in their ablation study on landmark localization that label adaptation is critical for benefiting from synthetic data. 

\section{Real-Windows Dataset}

\begin{figure*}[htb]
    \centering
    \def\svgwidth{\linewidth}
\begingroup%
  \makeatletter%
  \providecommand\color[2][]{%
    \errmessage{(Inkscape) Color is used for the text in Inkscape, but the package 'color.sty' is not loaded}%
    \renewcommand\color[2][]{}%
  }%
  \providecommand\transparent[1]{%
    \errmessage{(Inkscape) Transparency is used (non-zero) for the text in Inkscape, but the package 'transparent.sty' is not loaded}%
    \renewcommand\transparent[1]{}%
  }%
  \providecommand\rotatebox[2]{#2}%
  \newcommand*\fsize{\dimexpr\f@size pt\relax}%
  \newcommand*\lineheight[1]{\fontsize{\fsize}{#1\fsize}\selectfont}%
  \ifx\svgwidth\undefined%
    \setlength{\unitlength}{605.93077172bp}%
    \ifx\svgscale\undefined%
      \relax%
    \else%
      \setlength{\unitlength}{\unitlength * \real{\svgscale}}%
    \fi%
  \else%
    \setlength{\unitlength}{\svgwidth}%
  \fi%
  \global\let\svgwidth\undefined%
  \global\let\svgscale\undefined%
  \makeatother%
  \begin{picture}(1,0.16217516)%
    \lineheight{1}%
    \setlength\tabcolsep{0pt}%
    \put(0,0){\includegraphics[width=\unitlength,page=1]{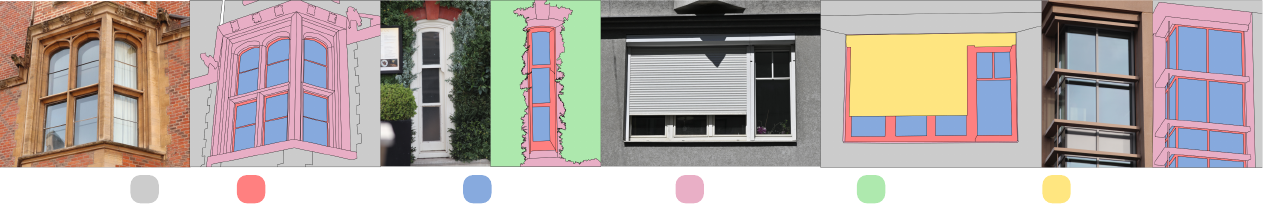}}%
    \put(0.12876332,0.00620547){\makebox(0,0)[lt]{\lineheight{1.25}\smash{\begin{tabular}[t]{l}wall\end{tabular}}}}%
    \put(0.21421645,0.00618807){\makebox(0,0)[lt]{\lineheight{1.25}\smash{\begin{tabular}[t]{l}window-frame\end{tabular}}}}%
    \put(0.39144971,0.00637118){\makebox(0,0)[lt]{\lineheight{1.25}\smash{\begin{tabular}[t]{l}window-pane\end{tabular}}}}%
    \put(0.56197443,0.0061122){\makebox(0,0)[lt]{\lineheight{1.25}\smash{\begin{tabular}[t]{l}wall-frame\end{tabular}}}}%
    \put(0.70436698,0.0061122){\makebox(0,0)[lt]{\lineheight{1.25}\smash{\begin{tabular}[t]{l}misc-object\end{tabular}}}}%
    \put(0.85324972,0.00807274){\makebox(0,0)[lt]{\lineheight{1.25}\smash{\begin{tabular}[t]{l}blind\end{tabular}}}}%
  \end{picture}%
\endgroup%

    \caption{Examples of the labels used to annotate our data. Each instance receives its own polygon. The reader may wish to zoom into the figure for details.  }
    \label{fig:labels}
\end{figure*}
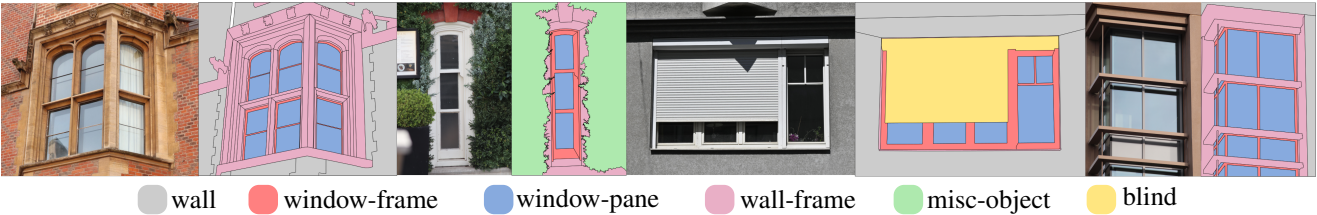

In line with our goal of advancing procedural modeling, our dataset is curated to offer high-resolution, diverse imagery essential for developing and evaluating procedural models.
Our dataset, comprising over \imgcount{} high-resolution images (up to $4K\times6K$), is one of the largest and most detailed collections of window imagery available. Unlike other datasets sourced from the web or Flickr~\cite{karras2019style,karras_progressive_celebahq}, we hold complete copyright ownership of every image, ensuring legal clarity and flexibility for research use\footnote{While the datasets we referenced use images under permissive licenses, owning our images outright simplifies usage rights.}.

The dataset's diversity in location, viewpoint, and architectural style, 
was achieved through a global effort. We engaged photographers from various countries, primarily hired via Upwork, to capture a wide array of window designs. 
This global collection effort ensures our dataset represents a broad spectrum of cultural and architectural diversity.
This diversity is crucial for procedural models to learn and adapt to a wide range of architectural styles, directly supporting our goal of creating versatile and realistic models.
Each image was chosen to highlight the window's design and architectural context, ensuring clear, distraction-free, and well-framed shots. The emphasis was on professional-grade photography with a focus on 4K resolution, achieving at least 2K pixels across each window, and taken during the day to ensure balanced exposure and minimal noise. 

In curating this collection, we also prioritized ethical considerations, instructing photographers to avoid capturing private situations or sensitive locations. Over 12 months, this project involved hiring 30 photographers and incurring costs of US \$0.20-\$0.50 per image, in addition to our quality control and subcontract management expenses.


The diversity of image locations is indicated in Table 1 of the Appendix\footnote{Available online \href{https://twak.github.io/winsyn/winsyn_appendix.pdf}{https://twak.github.io/winsyn/winsyn\_appendix.pdf}}, along with the number of images that include semantic segmentation labels and RAW camera data. The RAW camera data's higher bit-depth could be particularly valuable for future procedural modeling research, offering richer information for model training and evaluation. 

Many photos included multiple windows.  We manually annotated crops in each image as a region that includes a single window in the center, along with any portion of the wall that may have been adapted to the window (such as brickwork or molding) and a portion of the wall on all four sides of the window. 
Due to this cropping, window images are in a variety of sizes, as shown in Appendix Fig. 2. 
We store the original images and the cropping information separately and generate a cropped version of the dataset on demand. 

Each image has been carefully cropped and then annotated to focus on a single window and its immediate architectural context, as illustrated in Fig.~\ref{fig:labels}.
The annotations are designed to test if procedural models accurately replicate real-world architectural elements.
The cropped window images also underwent a manual annotation process for panoptic segmentation, and subsequent review for quality assurance, leading to the acquisition of \labelcount{} annotations at an average cost of US \$3.90 per image. While manageable, this cost notably exceeds image acquisition expenses, underscoring the value of methods that reduce reliance on labor-intensive labeled data for segmentation tasks.
A detailed account of the annotation process is presented in Section \ref{sec:semantic-segmentation}.  

The \name dataset serves as a foundational step towards our larger goal of procedural urban modeling, offering a comprehensive and detailed resource for both current and future research in this domain.

\section{Procedural Model Development}\label{sec:procedural-model-development}

\begin{table}
     \centering
\begin{tabular}{ccccc}
Label & Images Using & Area \% \\
\hline
wall & 8907 & 43.02\%\\
window pane & 8362 & 22.58\%\\
wall frame & 8697 & 14.91\%\\
window frame & 8681 & 9.71\%\\
unlabeled & 2994 & 3.09\%\\
shutter & 931 & 2.56\%\\
balcony & 973 & 1.08\%\\
misc object & 2357 & 1.07\%\\
blind & 375 & 0.75\%\\
bars & 679 & 0.68\%\\
open-window & 977 & 0.55\%\\
\hline
\end{tabular}

\caption{The labels used, their frequency of use, and percentage by area for the square-cropped dataset used for our experiments. We note that the dataset is a mix of well-used labels such as wall and less-used ones, such as blind or bar. The `unlabeled` category contained areas beyond the building (e.g., sky, streets) and a much smaller number of ambiguous areas where we could not reach a decision on how to label a feature.}
    \label{tab:labels-by-type}
\end{table}

In developing our procedural model, we aim to create a realistic synthetic dataset with the ability to produce variations (ablations and experiments) for evaluation against real-world data. The compact domain (windows are relatively simple to model) allows well-developed approaches to be used to create the scene geometry including Split Shape Grammars~\cite{wonka2003instant} and the CGA language~\cite{Mueller:2006:PMB}, combined with Bézier splines for curved window shapes. This approach ensures that our model can generate a wide range of architectural styles and window designs closely resembling real-world distributions. The canonical orientation offered by windows as a domain allows a layered approach to geometry generation, as Appendix Fig. 11: from the camera, we generate street clutter (e.g., cars, bollards, shrubs), wall decorations (shutters, balconies, pipes) exterior walls, window geometry, window dressing (blinds, curtains), and interior geometry. 

The procedural modeling pipeline, developed in Python, Blender~\cite{blender}, and rendered with the physically based Cycles~\cite{cycles} renderer is used to generate a diverse dataset of \syncount synthetic window images with corresponding labels (as Appendix Fig. 4). Normal, depth (Fig.~\ref{fig:overview}), and edge maps are optionally generated. The model was designed to prioritize diversity, making use of domain randomization to extend beyond typical real-world variations.

The procedural model's core uses two varieties of Split Grammar. The first, utilizing the CGA language, was employed to subdivide the volume to create the building mass, \facades, shutters, blinds, balconies, roof, and window-bounds. From these window-bounds, a second Split Shape Grammar creates the window shapes themselves, using B\`ezier spline curves to capture various geometries, such as trapezoid, arched, or circular windows. This grammar subdivides closed curves to create nested window frames and offsets them to create individual glass panes. The frame geometry is created by extruding a profile along each B\`ezier. The profiles are selected from a randomly selected hierarchy of profiles.
Parts of the window geometry hierarchy can be translated and rotated to `open' windows by sliding or hinged mechanisms.
This multi-grammar approach allows for detailed and varied window designs, while the generation remains algorithmically robust at scale. 

The procedural model is highly parameterized, with the number of parameters ranging from 216 to \paramcount in our dataset, depending on the chosen sequence of randomly-sampled rules in the grammars. We observe that optionally reusing parameters between parts of the model can improve visual realism; for example, we see window frames sharing their material (paint or stucco) with walls, or adjacent windows having similar (but not identical) materials.
In the analysis section below we vary the distributions of these parameters to ablate and experiment on our procedural model and identify the most performant factors, including materials, textures, geometry, camera position, and lighting.

Textures are primarily procedural shaders \cite{burley2012physically}, controlling materials such as wood, brick, or glass applied to appropriate object classes. To add realism, we captured exterior clutter, such as building signage, vehicles, and trash cans, through a number of sources, including pre-existing datasets and LiDAR/RGB scanning (Appendix Fig. 12) with bitmap textures. 

Scene lighting is supplied by a combination of skybox emission, sun-lamp, and optional interior sources. 
We use interior and exterior panoramic images for the skybox and interior-box to create realistic background environments for our geometry.
The camera is strategically positioned in order to capture the entire window from a predominantly frontal view; we use a variety of camera distances and adjust the field of view to frame the target window in the \facade.
Additional details of the procedural model can be found in Appendix Section 6.

\section{Estimating Procedural Model Quality} \label{sec:semantic-segmentation}

We use semantic segmentation as a benchmark to evaluate the quality of a procedural model. This method involves training segmentation models on labeled images generated from the procedural model, evaluating model performances on a holdout (\emph{test}) set of labeled real-world images, and providing a robust measure of the synthetic data's ability to accurately reflect real-world scenarios.

\begin{figure*}
    \centering
    \def\svgwidth{\linewidth}
    \begin{small}
    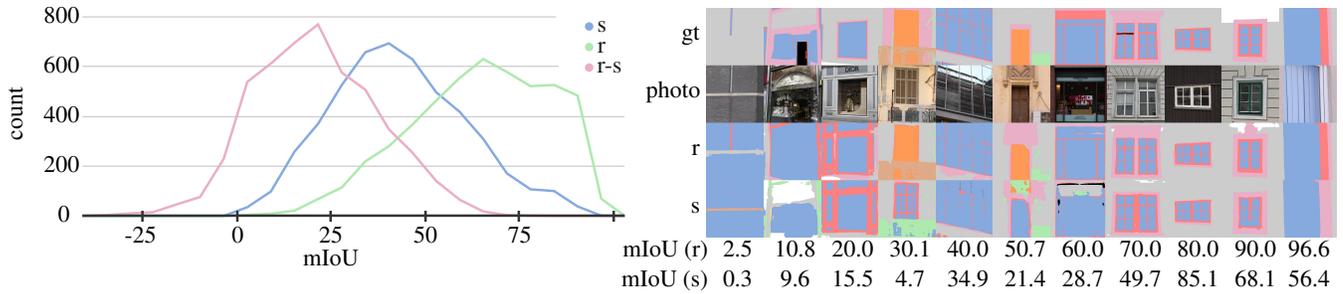
    \end{small}
    \caption{Left: histogram of per-image mIoUs showing the distribution of labeling results for a model trained on $n=$ 2,048 synthetic (s) and real (r) images. We also show the difference between the model mIoU's per image (r-s). The mIoU was evaluated on \testsize real images. Right: random samples of the labeling quality for networks trained on real and synthetic data; the first sample with an mIoU above each decile was selected. }
    \label{fig:by_miou}
\end{figure*}

For this approach to be successful, we must align our semantic labels with both real-world images and images generated by a procedural model. We established ten broad categories such as `window-pane', `window-frame', and `wall frame' (see Appendix Section 7.7). This categorization mirrors the elements that our procedural model can realistically generate. Furthermore, we label each unique instance within the images, as demonstrated in Fig.~\ref{fig:labels}, making the data suitable for panoptic segmentation even though we only evaluate the semantic segmentation task. This detailed set of labels extends traditional segmentation datasets, which typically provide a single `window' label, with occasional additions like `shutter' or `blind'~\cite{CMPFacade_Tylecek13,etrims-korc-forstner-tr09,LabelMeFacade-Brust15:ECP}. Our labeling allows consistent annotation across varied architectural styles but also ensures compatibility with the capabilities of procedural modeling.

When utilizing mIoU to evaluate procedural model performance, inspecting plots such as Figures~\ref{fig:render_spp} and~\ref{fig:lvl_1} can be insightful. These visual analyses aid in identifying parameters that optimize the model's performance. Additionally, Spearman's rank correlation is applied to quantify the impact of changes in model parameters (such as textures and lighting) on the synthetic image quality. 

\section{Analysis}

In all following experiments, we fine-tune a BEiT `base' model that was pre-trained on ImageNet-1k, trained on ImageNet-21k, and fine-tuned on data from some variation of our procedural model. We evaluated the mIoU over 10 labels, excluding the `unlabeled' category. All cropped window-images are resized to 512 pixels square; examples are shown in Fig.~\ref{fig:by_miou}.

The performance of our procedural model is best understood in the context of regional performance disparities within real-world data.
We contextualize the procedural model by comparing it to subsets from various regions of real-world data in Table~\ref{tab:cross-compare} shows. Each set consists of 1,024 training images and 300 test images. Models trained on synthetic data have a narrower performance range (29.33 to 35.02 mIoU) compared to those trained on locale-specific data (25.23 to 61.85 mIoU). Notably, synthetic data can surpass real data from a different locale, as seen in the comparison of England and Egypt results. This narrower range seems to indicate synthetic data has not over-fit any particular region.  The overall mIoU gap on the global dataset underscores the procedural model's limitations. The combined training set of real-world images evaluated to 53.79 mIoU, which is an upper limit on what one should expect from any model. The synthetic data yields 31.23 mIoU on the global test set, whereas other locales varied from 37.76--51.22 mIoU for the `other' set.  
Given the global set includes holdout images from each locale, it is not surprising the mIoU's are slightly higher. This table indicates that the procedural model is of comparable quality to choosing a single locale.


\begin{table*}
\centering

\begin{tabular}{l|l|lllllll}
& & \multicolumn{7}{c}{test} \\
\hline
 &  & global & Austria & Egypt & UK & USA & other & synthetic \\
\hline
 \parbox[t]{2mm}{\multirow{7}{*}{\rotatebox[origin=c]{90}{train}}} & global & {\cellcolor[HTML]{023858}} \color[HTML]{F1F1F1} 53.79 & {\cellcolor[HTML]{023858}} \color[HTML]{F1F1F1} 58.20 & {\cellcolor[HTML]{045B8F}} \color[HTML]{F1F1F1} 56.85 & {\cellcolor[HTML]{023858}} \color[HTML]{F1F1F1} 42.64 & {\cellcolor[HTML]{023858}} \color[HTML]{F1F1F1} 50.16 & {\cellcolor[HTML]{023858}} \color[HTML]{F1F1F1} 54.74 & {\cellcolor[HTML]{D4D4E8}} \color[HTML]{000000} 31.21 \\
 & Austria & {\cellcolor[HTML]{A9BFDC}} \color[HTML]{000000} 39.49 & {\cellcolor[HTML]{023858}} \color[HTML]{F1F1F1} 58.23 & {\cellcolor[HTML]{F2ECF5}} \color[HTML]{000000} 28.36 & {\cellcolor[HTML]{D9D8EA}} \color[HTML]{000000} 34.40 & {\cellcolor[HTML]{F1EBF4}} \color[HTML]{000000} 35.67 & {\cellcolor[HTML]{B9C6E0}} \color[HTML]{000000} 41.32 & {\cellcolor[HTML]{FFF7FB}} \color[HTML]{000000} 21.87 \\
 & Egypt & {\cellcolor[HTML]{1077B4}} \color[HTML]{F1F1F1} 47.51 & {\cellcolor[HTML]{BDC8E1}} \color[HTML]{000000} 38.32 & {\cellcolor[HTML]{023858}} \color[HTML]{F1F1F1} 61.85 & {\cellcolor[HTML]{F2ECF5}} \color[HTML]{000000} 33.02 & {\cellcolor[HTML]{D2D3E7}} \color[HTML]{000000} 38.01 & {\cellcolor[HTML]{0570B0}} \color[HTML]{F1F1F1} 49.75 & {\cellcolor[HTML]{D3D4E7}} \color[HTML]{000000} 31.34 \\
 & UK & {\cellcolor[HTML]{C2CBE2}} \color[HTML]{000000} 37.76 & {\cellcolor[HTML]{A1BBDA}} \color[HTML]{000000} 40.68 & {\cellcolor[HTML]{FFF7FB}} \color[HTML]{000000} 25.23 & {\cellcolor[HTML]{3D93C2}} \color[HTML]{F1F1F1} 38.56 & {\cellcolor[HTML]{F4EDF6}} \color[HTML]{000000} 35.39 & {\cellcolor[HTML]{EAE6F1}} \color[HTML]{000000} 37.64 & {\cellcolor[HTML]{E3E0EE}} \color[HTML]{000000} 28.51 \\
 & USA & {\cellcolor[HTML]{8EB3D5}} \color[HTML]{000000} 41.08 & {\cellcolor[HTML]{88B1D4}} \color[HTML]{000000} 42.50 & {\cellcolor[HTML]{F6EFF7}} \color[HTML]{000000} 27.39 & {\cellcolor[HTML]{EEE9F3}} \color[HTML]{000000} 33.30 & {\cellcolor[HTML]{02395A}} \color[HTML]{F1F1F1} 50.08 & {\cellcolor[HTML]{D7D6E9}} \color[HTML]{000000} 39.39 & {\cellcolor[HTML]{ECE7F2}} \color[HTML]{000000} 27.02 \\
 & other & {\cellcolor[HTML]{045788}} \color[HTML]{F1F1F1} 51.22 & {\cellcolor[HTML]{2786BB}} \color[HTML]{F1F1F1} 48.54 & {\cellcolor[HTML]{A1BBDA}} \color[HTML]{000000} 39.47 & {\cellcolor[HTML]{71A8CE}} \color[HTML]{F1F1F1} 37.44 & {\cellcolor[HTML]{89B1D4}} \color[HTML]{000000} 41.33 & {\cellcolor[HTML]{045382}} \color[HTML]{F1F1F1} 52.74 & {\cellcolor[HTML]{E4E1EF}} \color[HTML]{000000} 28.39 \\
 & synthetic & {\cellcolor[HTML]{FFF7FB}} \color[HTML]{000000} 31.23 & {\cellcolor[HTML]{FFF7FB}} \color[HTML]{000000} 29.53 & {\cellcolor[HTML]{EEE9F3}} \color[HTML]{000000} 29.33 & {\cellcolor[HTML]{FFF7FB}} \color[HTML]{000000} 32.15 & {\cellcolor[HTML]{FFF7FB}} \color[HTML]{000000} 34.17 & {\cellcolor[HTML]{FFF7FB}} \color[HTML]{000000} 35.02 & {\cellcolor[HTML]{023858}} \color[HTML]{F1F1F1} 62.12 \\
\end{tabular}
\caption{mIoU for different splits of the real labeled data on the segmentation task. Trained on 1,024, tested on 300 samples. \emph{global} is a mixture of all the real data; \emph{other} data is from locales outside of Austria, Egypt, UK, or USA. In this experiment our synthetic network is similarly trained on 1,024 samples from our baseline synthetic model.}
    \label{tab:cross-compare}
\end{table*}

\begin{figure}
    \centering
    \def\svgwidth{\linewidth}
    \begin{small}
    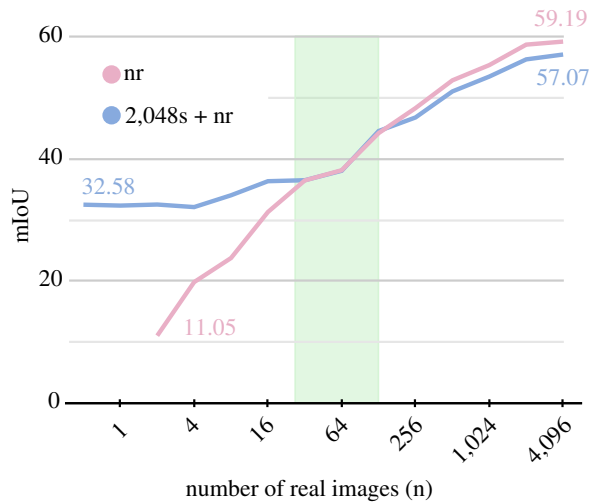
    \end{small}
    \caption{
          The effect of varying real-world dataset sizes on mIoU with (blue) and without (red) an additional 2,048 synthetic samples. The green area bounds the range in which adding real data neither harms nor improves performance; the right-most point of which has $n=152$ with an mIoU of $44.96$. At larger datasets, synthetic data slightly reduces mIoU relative to only using real-world data.
    }
    \label{fig:how-much-real-for-2k-synthetic}
\end{figure}

\subsection{Procedural Model Variations}

We assess the impact of synthetic dataset variations on a segmentation model by comparing against our baseline model. For each variation we render $2,048$ training examples and use the same geometry, lighting, and rendering settings as the baseline unless specified. For each variation, we render 2,048 training examples. Empirically, we find that the performance of synthetic data levels off at this number of examples (see Appendix Fig. 7) and evaluation is faster if the set is kept smaller. Evaluation takes place on a test partition of \testsize real-world images; the mIoU of the baseline model is \basemiou and real photos have an mIoU of \realmiou. To gauge each variation's importance, we either the relative mIoU range as a percentage of the baseline, or report the mIoU Spearman's rank correlation ($r_s$).  Detailed results are in Appendix Section 7 and summarized here. 

\begin{figure}
    \centering
    \def\svgwidth{1.05\linewidth}
    \begin{small}
    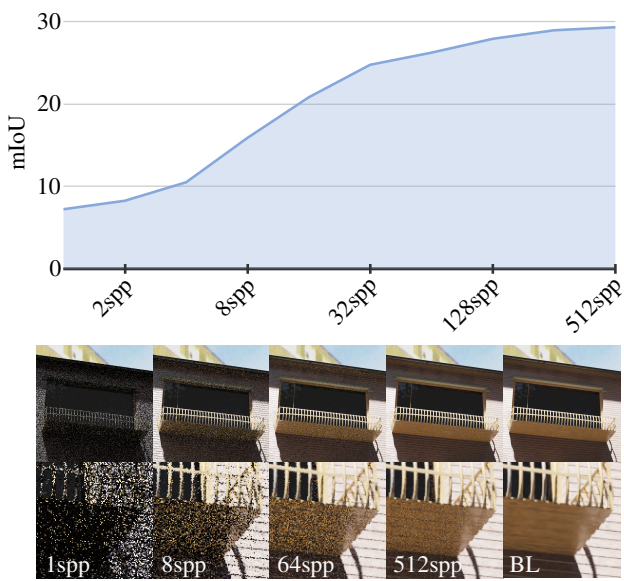
    \end{small}
    \caption{Top: The impact of rendering samples per pixel (spp) on segmentation task accuracy. Bottom: example renderings at different spp with zoomed section of rendering. BL = baseline.}
     \label{fig:render_spp}
\end{figure}

\textbf{Rendering samples.}  We evaluated the impact of samples per pixel (spp) on render quality, noting diminishing returns beyond 256spp (Fig. \ref{fig:render_spp}). Render times scale from 6.4s at 1spp to 85.2s at 512spp. A strong correlation  ($r_s=1,n=10$) exists between spp and mIoU, with a 68\% change in mIoU scores relative to baseline, underlining the importance of spp.  In this experiment no denoising was performed, however, our baseline model used 256spp and a powerful neural denoiser. 

\textbf{Mixed Materials.} We experimented with 9 material variations, including uniform gray, edge shaders, and random textures. Some tests used a single material for the entire scene, others assigned different materials per object. Simplifying materials, such as rendering only the albedo channel, resulted in a significant performance drop. Appendix Fig. 18 shows that any material restriction led to at least an 18\% mIoU decrease from the baseline. 

\textbf{Lighting.} We examined 8 lighting model variations and their mIoU impact, as detailed in Appendix Fig. 19. Lighting models like albedo-only were crucial, while lighting conditions such as night-time had moderate importance. Conditions varied mIoU by 15.35\% from baseline. Daylight-only training decreased baseline mIoU by 1.04\% relative to baseline.

\textbf{Camera.} We experimented with the distribution of camera positions. These 7 variations used a simple model which sampled a camera position over a circle, of radius $r=\{0...96\}$ meters, truncated at the floor plane. The circle is positioned $5$ meters from the wall, directly in front of the window. The camera's field of view is adjusted to the apparent window size. We observed limited impact on mIoU as $r$ changes; peak task performance was at $r=12m$.

\textbf{Window Geometry.} We ran 7 tests with varying window dimensions and shapes, including square and non-rectangular windows. The mIoU impact was minor, fluctuating by up to 5.7\% relative to the baseline (1.86 absolute), with the best variation 1.2\% worse relative to the baseline ($-0.04$ absolute). Small features, though noticeable to humans, had a weak impact on model performance. 

\begin{figure}
    \centering
    \def\svgwidth{\linewidth}
    \begin{small}
    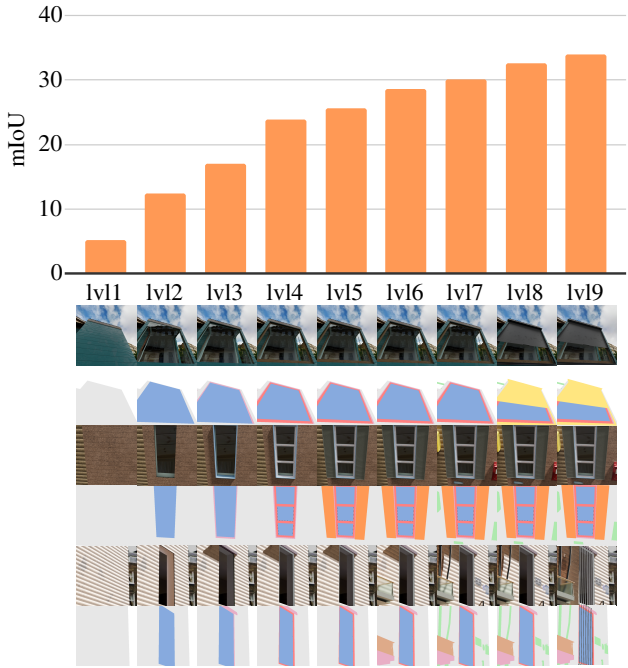
    \end{small}
    \caption{Top: The progressive impact of incorporating additional labels into the procedural model, with label subsets expanding from lvl1 to lvl9. The sequence begins with `walls' only (lvl1), adding `window panes' (lvl2), `wall frames' (lvl3), `window frames' (lvl4), `shutters' (lvl5), `balconies' (lvl6), `miscellaneous objects' (lvl7), `blinds' (lvl8), and `bars' (lvl9). The full model includes additional features beyond lvl9, such as interior dressing, open windows, and windows without glass. Bottom: Examples corresponding to each incremental level of model complexity.}
    \label{fig:lvl_1}
\end{figure}

\textbf{Labels Modeled.} In developing our procedural synthetic data generator, we prioritized labels by size, starting with \emph{wall} and ending with \emph{open-window} (Appendix Section 4). This enabled the assessment of mIoU at nine developmental stages (Fig.~\ref{fig:lvl_1}). Adding smaller classes later showed diminishing returns and occasionally reduced single-label accuracy (Appendix Fig. 22).

\subsection{Additional Experiments}

In our segmentation experiments, we employed the BEiTv2 base model~\cite{beit2,bao2022beit} for its proficiency in generating accurate results with minimal data. However, the trends we observed are consistent across various models, including DeepLabv3+~\cite{deeplabv3p}.
In the Appendix Section 5, we discuss two experiments that study the impact on the segmentation task when training (and testing) with different mixes of real and synthetic data. The first experiment in Appendix Fig. 7 explores segmentation performance when training and testing on different amounts of real or synthetic data. We observe that synthetic data saturates (stops improving with additional data), while real-world data continues to benefit from additional samples. The second experiment, in Appendix Fig. 8, illustrates the impact of different mixtures of real and synthetic data, when tested on real data. Mixing synthetic and real data can be advantageous when there is little real data, but adding large amounts of synthetic data does not help task performance. We conclude that no volume of synthetic data can overcome this domain gap.

An exploration of different architectures and data partitions is presented in Appendix Fig. 9 and Table 2 - we train an older convolution network without pre-training (DeepLab3+~\cite{deeplabv3p}), a `large` BEiTv2 model, an architecture which uses real data labels during training to improve performance (Label Adaptation~\cite{wood_fake_2021}), another technique which uses unlabeled data to improve performance (MIC~\cite{hoyer2023mic}), adjusting the colors of the training dataset (Histogram Matching~\cite{gonzales1987digital}), and creating an `easy' real data partition. Under these approaches, the synthetic-real domain gap persists, suggesting that network architecture and data partitions are an orthogonal research direction to improving the quality of synthetic procedural models.

\section{Acknowledgements}

We thank our lead photographers Michaela Nömayr and Florian Rist, and engineer Prem Chedella, as well as our contributing photographers: Aleksandr Aleshkin, Angela Markoska, Artur Oliveira, Brian Benton, Chris West, Christopher Byrne, Elsayed Saafan, Florian Rist, George Iliin, Ignacio De Barrio, Jan Cuales, Kaitlyn Jackson, Kalina Mondzholovska, Kubra Ayse Guzel, Lukas Bornheim, Maria Jose Balestena, Michaela Nömayr, Mihai-Alexandru Filoneanu, Mokhtari Sid Ahmed Salim, Mussa Ubapa, Nestor Angulo Caballero, Nicklaus Suarez, Peter Fountain, Prem Chedella, Samantha Martucci, Sarabjot Singh, Scarlette Li, Serhii Malov, Simon R. A. Kelly, Stephanie Foden, Surafel Sunara, Tadiyos Shala, Susana Gomez, Vasileios Notis, Yuan Yuan, and finally labelyourdata.com for the labeling.

\section{Conclusion}

We have introduced a new dataset of \imgcount photos (2.09 terapixels), 
 of which \labelcount photos are semantically labeled, as well as a high-quality procedural model that closely approximates real-world variation, making it effective for image segmentation tasks. Together these components of \name have applications in unsupervised domain adaptation (using the large amount of unlabeled data), super-resolution (via the thigh resolution images), learning from RAW sources, generative modeling, 3D reconstruction or depth recovery (using the synthetic depth or 3D geometry from the procedural model), as well as learning from images containing transparent or specular materials (i.e., window panes/glass balconies are often transparent).

We systematically explored the effect of variations on our model on mIoU over \variationcount variations and \totalrenders renders. The mIoU performance gap between our synthetic and real-world data is comparable to inter-country differences with largely different architectural styles. However, the difference between synthetic data and real data is still much larger than desired. This gap is not sufficiently reduced by either our work or other network architectures or dataset scale. We, therefore, believe that \name provides a timely and efficient testbench from which researchers can iterate quickly to explore graphics contributions to the synthetic data problem. Our work contributes a sizable and versatile dataset that can be the basis of exciting and much-needed progress in the area of procedural graphics for synthetic data generation in machine learning.
The key value of our dataset is that as a new benchmark of simulated and real images, it enables others to study the problem at an approachable level of complexity.

{
    \small
    \bibliographystyle{ieeenat_fullname}
    \bibliography{references}
}


\end{document}